\documentclass{article} 
\usepackage{iclr2020_conference,times}

\usepackage{amsmath,amsfonts,bm}

\def\eqref#1{equation~\ref{#1}}

\def\1{\bm{1}}

\DeclareMathAlphabet{\mathsfit}{\encodingdefault}{\sfdefault}{m}{sl}
\SetMathAlphabet{\mathsfit}{bold}{\encodingdefault}{\sfdefault}{bx}{n}

\newcommand{\E}{\mathbb{E}}

\DeclareMathOperator*{\argmax}{arg\,max}

\DeclareMathOperator{\Tr}{Tr}

\usepackage{algorithm,algpseudocode}
\usepackage{hyperref}
\usepackage{url}
\usepackage{times}
\usepackage{epsfig}
\usepackage{graphicx}
\usepackage{amsmath}
\usepackage{amssymb}
\usepackage{authblk}

\usepackage{booktabs}
\usepackage{multirow}
\usepackage{float}
\usepackage{bbm}
\usepackage{relsize}

\begin{document}

\title{Small-GAN: Speeding up GAN Training using Core-Sets}


\author[1,4]{\bfseries Samarth Sinha}
\author[2]{\bfseries Han Zhang}
\author[1]{\bfseries Anirudh Goyal}
\author[1,$\dagger$]{\bfseries Yoshua Bengio}
\author[2]{\bfseries Hugo Larochelle}
\author[2]{\bfseries Augustus Odena}
\affil[1]{Mila, Universit\'e de Montr\'eal} 
\affil[2]{Google Brain}
\affil[$\dagger$]{CIFAR Senior Fellow}
\affil[4]{University of Toronto}

\affil[ ]{\protect \\ \texttt{samarth.sinha@mail.utoronto.ca, augustusodena@google.com} }

%

\newcommand{\fix}{\marginpar{FIX}}
\newcommand{\new}{\marginpar{NEW}}

\iclrfinalcopy 

\maketitle

\begin{abstract}
Recent work by \citet{biggan} suggests that Generative Adversarial Networks (GANs) benefit disproportionately from large mini-batch sizes.
Unfortunately, using large batches is slow and expensive on conventional hardware.
Thus, it would be nice if we could generate batches that were {\it effectively large} though actually small.
In this work, we propose a method to do this, inspired by the use of Coreset-selection in active learning.
When training a GAN, we draw a large batch of samples from the prior and then compress that batch using Coreset-selection.
To create effectively large batches of `real' images, we create a cached dataset of Inception activations of each training image, 
randomly project them down to a smaller dimension, and then use Coreset-selection on those projected activations at training time.
We conduct experiments showing that this technique substantially reduces training time and memory usage for modern GAN variants, 
that it reduces the fraction of dropped modes in a synthetic dataset, and that it allows GANs to reach a new state of the art in anomaly detection.
\end{abstract}

\section{Introduction}

Generative Adversarial Networks (GANs)~\citep{gan} have become a popular research topic. 
Arguably the most impressive results have been in image synthesis \citep{biggan, otgan, sngan,sagan,stackgan,CRGAN}, but they have also been applied fruitfully to text generation \citep{maskgan, leakgan}, domain transfer learning \citep{cyclegan, stackgan, pix2pix}, and various other tasks \citep{xian2018feature, ledig2017photo, zhu2017generative,GANFIGHT}.

Recently, \citet{biggan} substantially improved the results of \citet{sagan} by using
very large mini-batches during training.
The effect of large mini-batches in the context of deep learning is well-studied \citep{smith2017don, goyal2017accurate, keskar2016large, BIGBATCH} and general consensus
is that they can be helpful in many circumstances, but the results of \citet{biggan} suggest that GANs benefit disproportionately from large batches \citep{OPENQUESTIONS}.
In fact, Table 1 of \citet{biggan} shows that for the Frechet Inception Distance (FID) metric \citep{FID}
on the ImageNet dataset, scores can be improved from 18.65 to 12.39 simply by making the batch eight times larger.

Unfortunately, increasing the batch size in this manner is not always possible since it
increases the computational resources required to train these models -- 
often beyond the reach of conventional hardware. The experiments from the BigGAN 
paper require a full `TPU Pod'. The `unofficial' open source release of BigGAN 
works around this by accumulating gradients across 8 different V100 GPUs and only 
taking an optimizer step every 8 gradient accumulation steps.
Future research on GANs would be much easier if we could have the gains from large
batches without these pain points.
In this paper, we take steps toward accomplishing that goal by proposing a technique
that allows for \textit{mimicking} large batches without the computational costs
of actually using large batches.

In this work, we use Core-set selection \citep{agarwal2005geometric} to sub-sample
a large batch to produce a smaller batch.
The large batches are then discarded, and the sub-sampled, smaller, batches are used to train the GAN.
Informally, this procedure yields small batches with `coverage' similar to 
that of the large batch -- in particular the small batch tries to `cover' all the same
modes as are covered in the large batch.
This technique yields many of the benefits of having large batches with much less computational overhead.
Moreover, it is generic, and so can be applied to nearly all GAN variants.

Our contributions can be summarized as follows:

\begin{itemize}
    \item We introduce a simple, computationally cheap method to increase the 
    `effective batch size' of GANs, which can be applied to any GAN variant.
    \item We conduct experiments on the CIFAR and LSUN datasets showing 
    that our method can substantially improve FID across different GAN 
    architectures given a fixed batch size.
    \item We use our method to improve the performance of the technique from 
    \citet{kumar2019maximum}, resulting in state-of-the-art performance at 
    GAN-based anomaly detection.
\end{itemize}

\section{Background and Notation}
\paragraph{Generative Adversarial Networks}
A Generative Adversarial Network (or GAN) is a system of two neural networks trained `adversarially'.
The generator, $G$, takes as input samples from a prior $z \sim p(z)$ and outputs the learned distribution, $G(z)$.
The discriminator, $D$, receives as input both the training examples, $X$, and the
synthesized samples, $G(z)$, and outputs a distribution
$D(.)$ over the possible sample source.
The discriminator is then trained to maximize the following objective:
\begin{equation}
\label{gen_eqn}
\mathcal{L}_D = -\E_{x \sim p_\text{data}} [\log D(x)] -
\E_{z \sim p(z)} [\log (1 - D(G(z)))]
\end{equation}
while the generator is trained to minimize\footnote{
This is the commonly used ``Non-Saturating Cost''.
There are many others, but for brevity and since our technique 
we describe is agnostic to the loss function, we will omit them.
}:
\begin{equation}
\label{dsc_eqn}
\mathcal{L}_G = -\E_{z \sim p(z)} [\log D(G(z))]
\end{equation}
Informally, the generator is trained to \textit{trick} the discriminator into believing that 
the generated samples $G(z)$ actually come from the target distribution, $p(x)$, while the
discriminator is trained to be able to distinguish the samples from each other.

\paragraph{Inception Score and Frechet Inception Distance:}
We will refer frequently to the Frechet Inception Distance (FID) \citep{FID}, 
to measure the effectiveness of an image synthesis model.
To compute this distance, one assumes that we have a pre-trained Inception 
classifier. 
One further assumes that the activations in the penultimate layer
of this classifier come from a multivariate Gaussian.
If the activations on the real data are $N(m, C)$ and the activations on the
fake data are $N(m_w, C_w)$, the FID is defined as:

\begin{equation}
\|m-m_w\|_2^2+ \Tr \bigl(C+C_w-2\bigl(CC_w\bigr)^{1/2}\big)
\end{equation}

\paragraph{Core-set selection:}

In computational geometry, a Core-set, $Q$, of a set $P$ is a subset $Q \subset P$ that 
approximates the `shape' of $P$ \citep{agarwal2005geometric}.
Core-sets are used to quickly generate approximate solutions to problems whose full solution on the original set would be burdensome to compute.
Given such a problem\footnote{
As an example, consider computing the diameter of a point-set \citep{agarwal2005geometric}.}, one computes $Q$,
then quickly computes the solution to the problem for $Q$ and converts that into an approximate solution for the original set $P$.
The general Core-set selection problem can be formulated several ways, here we consider the 
the \textit{minimax facility location} formulation \citep{farahani2009facility}:

\begin{equation}
    \min_{Q : |Q| = k} \max_{x_i \in P} \min_{x_j \in Q} d(x_{i}, x_{j})
\end{equation}

where $k$ is the desired size of $Q$, and $d(., .)$ is a metric on $P$. 
Informally, the formula above encodes the following objective: Find some set, $Q$, of points of size $k$ such that the maximum distance between a point in $P$ and its nearest point in $Q$ is minimized.
Since finding the exact solution to the minimax facility location problem is NP-Hard \citep{wolsey2014integer}, we will have to make do with a greedy approximation, detailed
in Section \ref{section:greedy}.

\begin{algorithm}
    \caption{\texttt{GreedyCoreset}} 
    \label{alg:coreset}
\begin{algorithmic}
        \renewcommand{\algorithmicrequire}{\textbf{Input:}}
        \renewcommand{\algorithmicensure}{\textbf{Output:}}
        \Require batch size ($k$), data points ($x$ where $|x| > k$)
        \Ensure subset of $x$ of size $k$
        \State $s \gets \{\}$  \Comment{Initialize the sampled set}
        \While{$|s| < k$}  \Comment{Iteratively add points to sampled set}
            \State $p \gets \argmax_{x_i \notin s} \min_{x_j \in s} d(x_{i}, x_{j})$
            \State $s \gets s \cup \{p\}$
        \EndWhile
        \Return $s$
\end{algorithmic}
\end{algorithm}

\section{Using Core-set Sampling for GANs (or Small-GAN)}
We aim to use Core-set sampling to increase the effective batch size during GAN training.
This involves replacing the basic sampling operation that is done implicitly when 
minibatches are created.
This implicit sampling operation happens in two places:
First, when we create a minibatch of samples drawn from the prior distribution $p(z)$.
Second, when we create a minibatch of samples from the target distribution $p_{\texttt{data}}(x)$ to update the parameters of the discriminator.
The first of these replacements is relatively simple, while the second presents challenges.
In both cases, we have to work around the fact that actually doing Core-set sampling is 
computationally hard.

\subsection{Sampling from the Prior}
We need to sample from the prior when we update the discriminator and generator parameters.
Our Core-set sampling algorithm doesn't take into account the geometry of the space we 
sample from, so sampling from a complicated density might cause trouble. 
This problem is not intractable, but it's nicer not to have to deal with it, so in 
the absence of any evidence that the form of the prior matters very much, 
we define the prior in our experiments to be the uniform distribution over a hypercube. 
To add Core-set sampling to the prior distribution, we sample $n$ points from the prior, where $n$ is greater than the desired batch size, $k$. 
We then perform Core-set selection on the large batch of size $n$ to create a batch of 
size $k$. 
The smaller batch is what's actually used to perform an SGD step.

\subsection{Sampling from the Target Distribution}
Sampling from the target distribution is more challenging.
The elements drawn from the distribution are high dimensional images, so taking
pairwise distances between them will tend to work poorly due to concentration of distances \citep{donoho2000high, vaal}, and the fact that Euclidean distances are semantically meaningless in image space \citep{girod1993s, eskicioglu1995image}.

To avoid these issues, we instead pre-process our data set by computing the `Inception
Embeddings' of each image using a pre-trained classifier \citep{szegedy2017inception}.
This is commonly done in the transfer-learning literature, where it is generally accepted
that these embeddings have nontrivial semantics \citep{yosinski2014transferable}.
Since this pre-processing happens only once at the beginning of training, it doesn't 
affect the per-training-step performance.

In order to further reduce the time taken by the Core-set selection procedure, and inspired
by the Johnson-Lindenstrauss Lemma \citep{dasgupta2003elementary}, we take random low dimensional projections of the Inception Embeddings.
Combined with Core-set selection, this gives us low-dimensional representations of the training set images in which pairwise Euclidean distances have meaningful semantics.
We can then use Core-set sampling on those representations to select images at training time, analogous to how we select images from the prior.

\subsection{Greedy Core-set Selection}
\label{section:greedy}
In the above sections, we have invoked Core-set selection while glossing over the detail
that exactly solving the $k$-center problem is NP-hard.
This is important, because we propose to use Core-set selection at \textit{every} training
step\footnote{
Though the Core-set sampling does happens on CPU and so could be done in parallel to the GPU operations used to train the model, as long as the Core-set sampling time doesn't exceed the time of a forward and backward pass -- which it doesn't.}.
Fortunately, we can make do with an approximate solution, which is faster to compute:
we use the greedy $k$-center algorithm (similar to \citet{sener2017active}) 
summarized in Alg. \ref{alg:coreset}.

\subsection{Small-GAN}
Our full proposed algorithm for GAN training is presented in Alg. \ref{alg:coreset_gan}. 
Our technique is agnostic to the underlying GAN framework and therefore can replace random sampling of mini-batches for all GAN variants. 
More implementation details and design choices are presented in Section \ref{exp}.

\begin{algorithm}
    \caption{\texttt{Small-GAN}} 
    \label{alg:coreset_gan}
\begin{algorithmic}
        \renewcommand{\algorithmicrequire}{\textbf{Input:}}
        \renewcommand{\algorithmicensure}{\textbf{Output:}}
        \Require target batch size ($k$), starting batch size ($n > k$), Inception embeddings ($\phi_{I}$)
        \Ensure a trained GAN
        \State Initialize networks $G$ and $D$
        \For{$step = 1 \text{ to } ...$}
            \State $z \sim p(z)$ \Comment{Sample $n$ points from the prior}
            \State $x \sim p(x)$ \Comment{Sample $n$ points from the data distribution}
            \State $\phi(x) \gets \phi_{I}(x)$ \Comment{Get cached embeddings for $x$}
            \State $\hat{z} \gets \texttt{GreedyCoreset}(z)$ \Comment{Get Core-set of $z$}
            \State $\widehat{\phi(x)} \gets \texttt{GreedyCoreset}(\phi(x))$ \Comment{Get Core-set of embeddings}
            \State $\hat{x} \gets \phi_{I}^{-1}(\widehat{\phi(x)})$  \Comment{Get $x$ corresponding to sampled embeddings}
            \State Update GAN parameters as usual
        \EndFor
\end{algorithmic}
\end{algorithm}

\section{Experiments}
\label{exp}

In this section we look at the performance of our proposed sampling method on various tasks:
In the first experiment, we train a GAN on a Gaussian mixture dataset with a large number of modes and confirm our method substantially mitigates `mode-dropping'.
In the second, we apply our technique to GAN-based anomaly detection \citep{kumar2019maximum} and significantly improve on prior results.
Finally, we test our method on standard image synthesis benchmarks and confirm that 
our technique seriously reduces the need for large mini-batches in GAN training.
The variety of settings in these experiments testifies to the generality of our proposed
technique.

\subsection{Implementation Details}

For our Core-set algorithm, the distance function, $d(\cdot,\cdot)$ s the $\ell_2$-norm for both the prior and target distributions. 
The hyper-parameters used in each experiment are the same as originally proposed in the paper introducing that experiment, unless stated otherwise.
For over-sampling, we use a factor of 4 for the prior $p(z)$ and a factor of 8 for the target, $p(x)$, 
unless otherwise stated.
We investigate the effects of different over-sampling factors in the ablation study in Section \ref{ablation}.

\subsection{Mixture of Gaussians}
\label{mog_section}
We first investigate the problem of mode dropping \citep{arora2018gans} in GANs, where the GAN generator is unable to recover some modes from the target data set.
We investigate the performance of training a GAN to recover a different number 
of modes of 2D isotropic Gaussian distributions, with a standard deviation of 0.05. We use 
a similar experimental setup as \citet{DRS}, where our generator and discriminator are parameterized
using 4 ReLU-fully connected networks, and use the standard GAN loss in Eq.\ \ref{gen_eqn} and \ref{dsc_eqn}. 
To evaluate the performance of the models, we generate $10,000$ samples and assign them to their closest
mode. 
As in \citet{DRS}, the metrics we use to evaluate performance are: $i)$ `high quality samples', which are samples within 4 standard deviations of the assigned mode and $ii)$ `recovered modes' which are mixture components with at least one assigned sample. 

Our results are present in table \ref{tab:gaussians_exp}, where we experiment with an increasing number
of modes. 
We see that as the number of modes increases, a normal GAN suffers from increased
mode dropping and lower sample quality compared to Core-set selection. 
With 100 modes, Core-set selection recovers 97.33\% of the modes compared to 90.67\%
for the vanilla GAN. Core-set selection also generates 49.87\% `high quality' samples
compared to 23.31\% for the vanilla GAN.

\begin{table}[t]
    \centering
    \begin{tabular}{|c|@{\hskip 0.05in}|c|c||c|c|}
    \hline
    Number of & \% of Recovered  & \% of Recovered & \% of High-Quality & \% of High-Quality\\
    Modes & Modes (GAN) & Modes (Ours) & Samples (GAN) & Samples (Ours) \\
     \hline 
     \hline
    25 & \textbf{100} & \textbf{100} & 95.76 & \textbf{98.9} \\ 
   
    \hline
    36 & \textbf{100} & \textbf{100} & 92.73 & \textbf{95.34} \\ 
    \hline
    49 & 98.12 & \textbf{99.85} & 84.28 & \textbf{88.1} \\ 
    \hline
    64 & 96.13 & \textbf{99.01} & 68.81 & \textbf{82.11} \\ 
    \hline
    81 & 92.59 & \textbf{98.84} &49.74 & \textbf{71.75} \\ 
    \hline
    100 & 90.67 & \textbf{97.33} & 23.31 & \textbf{49.87} \\ 
    \hline
    \end{tabular}
    \vskip 0.05in
    \caption{Experiments with large number of modes}
    \label{tab:gaussians_exp}
\end{table}

\subsection{Anomaly Detection}
 
\begin{table}[t]
    \centering
    \begin{tabular}{|c|@{\hskip 0.05in}|c|c|c|}
        \hline
        Held-out Digit & Bi-GAN & MEG & Core-set+MEG \\
        \hline
        \hline
        1 & 0.287 & 0.281 & \textbf{0.351} \\
        \hline
        4 & 0.443 & 0.401 & \textbf{0.501} \\
        \hline
        5 & 0.514 & 0.402 & \textbf{0.518} \\
        \hline
        7 & 0.347 & 0.29 & \textbf{0.387} \\
        \hline
        9 & 0.307 & 0.342 & \textbf{0.39} \\
        \hline
    \end{tabular}
    \caption{Experiments with Anomaly Detection on MNIST dataset. The Held-out digit represents the digit
    that was held out of the training set during training and treated as the anomaly class. The numbers
    reported is the area under the precision-recall curve.}
    \label{tab:anomaly_exp}
\end{table}
 
To see whether our method can be useful for more than just GANs, we also apply it to the 
Maximum Entropy Generator (MEG) from \citet{kumar2019maximum}.
MEG is an energy-based model whose training procedure requires maximizing the entropy of the 
samples generated from the model.
Since MEG gives density estimates for arbitrary data points, it can be used for 
anomaly detection -- a fundamental goal of machine learning research \citep{chandola2009anomaly,kwon2017survey} -- 
in which one aims to find samples that are `atypical' given a source data set.
\citet{kumar2019maximum} do use MEG successfully for this purpose, achieving results close to the state-of-the-art
technique for GAN-based anomaly detection \citep{bigan}.
We hypothesized that -- since energy estimates can in theory be improved by larger batch sizes -- these results 
could be further improved by using Core-set selection, and we ran an experiment to confirm this hypothesis.

We follow the experimental set-up from \citet{kumar2019maximum} by training the MEG 
with all samples from a chosen MNIST digit left-out during training. 
Those samples then serve as the `anomaly class' during evaluation.
We report the area under the precision-recall curve and average the score over the last 10 epochs.
The results are reported in Table \ref{tab:anomaly_exp}, which provides clear evidence in favor
of our above hypothesis: for all digits tested, adding Core-set selection to MEG substantially 
improves the results.
By performing these experiments, we aim to show the general applicability of Core-set selection,
not to suggest that MEG is superior to BiGANs \citep{bigan} on the task.
We think it's likely that similar improvements could be achieved by using Core-set selection with BiGANs.

\begin{table}[t]
    \centering
    \begin{tabular}{|c|c||c|c||c|c|}
        \hline
         &  Small-GAN &  & Small-GAN & & Small-GAN \\
        GAN (batch-  & (batch-size & GAN (batch- & (batch-size & GAN (batch- & (batch-size \\
        size = 128) & = 128) & size = 256) & = 256) & size = 512) & = 512)\\
        \hline
        \hline
        18.75 $\pm$ 0.2 & \textbf{16.73 $\pm$ 0.1} & 17.9 $\pm$ 0.1 & \textbf{16.22 $\pm$ 0.3} & 15.68 $\pm$ 0.2 & \textbf{15.08 $\pm$ 0.1} \\
        \hline
    \end{tabular}
    \caption{FID scores for CIFAR using SN-GAN as the batch-size is progressively doubled.
    The FID score is calculated using $50,000$ generated samples from the generator.}
    \label{tab:cifar_exp}
\end{table}

\begin{table}[t]
    \centering
    \begin{tabular}{|c||c|c|c|}
        \hline
        Small-GAN (batch- & GAN (batch- & GAN (batch- & GAN (batch-\\
        size = 64) & size = 64) & size = 128) & size = 256) \\
        \hline
        \hline
        13.08 & 14.82 & 13.02 & 12.63 \\
        \hline
    \end{tabular}
    \caption{FID scores for LSUN using SAGAN as the batch-size is progressively doubled.
    The FID score is calculated using $50,000$ generated samples from the generator.
    All experiments were run on the `outdoor church' subset of the dataset.}
    \label{tab:lsun_exp}
\end{table}

\subsection{Image Synthesis}

\paragraph{CIFAR and LSUN:}
We also conduct experiments on standard image synthesis benchmarks.
To further show the generality of our method, 
we experiment with two different GAN 
architectures and two image datasets. 
We use Spectral Normalization-GAN \citep{sngan} and Self Attention-GAN 
\citep{sagan} on the CIFAR \citep{cifar} and LSUN \citep{lsun} datasets, respectively.
For the LSUN dataset, which consists of 10 different categories, we train the model using the `outdoor church' subset of the data.

For evaluation, we measured the FID scores \citep{FID} of $50,000$ generated samples from
the trained models\footnote{Note that we measure the performance of all the models using the PyTorch
version of FID scores, and not the official Tensorflow one. We ran all our experiments 
with the same code for accurate comparison.}.
We compare the performance using SN-GANs with and without Core-set selection across progressively
doubling batch sizes. 
We observe a similar effect to \citet{biggan}: just by increasing the mini-batch size by a factor of 4, 
from 128 to 512, we are able to improve the FID scores from 18.75 to 15.68 for SN-GANs.
This further demonstrates the importance of large mini-batches for GAN training.
Adding Core-set selection significantly improves the performance of the underlying GAN 
for all batch-sizes.
For a batch size of 128, our model using Core-set sampling significantly outperforms the 
normal SN-GAN trained with a batch size of 256, and is comparable to an SN-GAN trained 
with a batch size of 512.
The results suggest that the models perform significantly better for any given batch size 
when Coreset-sampling is used.

However, Core-set sampling does become less helpful as the underlying batch size increases:
for SN-GAN, the performance improvement at a batch size of 128 is much larger than the
improvement at a batch size of 512.
This supports the hypothesis that Core-set selection works by approximating the coverage
of a larger batch; a larger batch can already recover more modes of the data - so under
this hypothesis, we would expect Core-set selection to help less.

We see similar results when experimenting with 
Self Attention GANs (SAGAN) \citep{sagan} on the LSUN dataset \citep{lsun}. 
Compared to our results with SN-GAN, increasing the batch size results in a smaller difference in the performance for the SAGAN model, but we still see the FID improve from 14.82 to 12.63 as the batch-size is increased by a factor of 4.
Using Core-set sampling with a batch size of 64, we are able to achieve a comparable score 
to when the model is trained with a batch size of 128.
We believe that one reason for a comparably smaller advantage of using Core-set sampling on
LSUN is the nature of the data itself:
using the `outdoor church' subset of LSUN reduces the total number of `modes'
\textit{possible} in the target distribution, since images of churches have fewer differences than the images in CIFAR-10 data set.
We see similar effects in the mixture of Gaussians experiment (See \ref{mog_section})
where the relative difference between a GAN trained with and without Core-set sampling
increases as the number of modes are increased.

\paragraph{ImageNet:}
Finally, in order to test that our method would work `at-scale', we ran an experiment on the ImageNet data set.
Using the code at \url{https://github.com/heykeetae/Self-Attention-GAN}, we trained two GANs:
The first is trained exactly as described in the open-source code.
The second is trained using Coreset selection, with all other hyper-parameters unchanged.
Simply adding Coreset selection to the existing SAGAN code materially improved the FID
(which we compute using 50000 samples): the baseline model had an FID of 19.40 and the
Core-set model had an FID of 17.33.

\subsection{Timing Analysis}
\begin{table}[t]
    \centering
    \begin{tabular}{|c|c|c|c|}
        \hline
        Small-GAN (batch & SN-GAN (batch & SN-GAN (batch & SN-GAN (batch \\
        size = 128) & size = 128) & size = 256) & size = 512) \\
        \hline
        14.51 & 13.31 & 26.46 & 51.64 \\
        \hline
    \end{tabular}
    \caption{Timing to perform 50 gradient updates for SN-GAN with and without Core-sets.
    The time is measured in seconds.
    All the experiments were performed on a single NVIDIA Titan-XP GPU.
    The sampling factor was 4 for the prior and 8 for the target distribution.}
    \label{tab:time_table}
\end{table}

Since random sampling can be done very quickly, it is important to investigate the amount
of time it takes to train GANs with and without Core-set sampling.
We measured the time for SN-GAN to do 50 gradient steps on the CIFAR dataset with various mini-batch sizes: the results are in Table \ref{tab:time_table}.
On average, for each gradient step, the time added by performing Core-Set sampling is only 0.024 seconds. 

\subsection{Ablation Study}
\label{ablation}

We conduct an ablation study to investigate the reasons for the effectiveness of Core-set selection.
We also investigate the effect of different sampling factors and other hyper-parameters.
We run all ablation experiments on the task of image synthesis using SN-GAN \citep{sngan} with the CIFAR-10 dataset \citep{cifar}.
We use the same hyperparameters as in our main image synthesis experiments and a batch size of 128, unless otherwise stated.

\subsection{Examination of Main Hyper-Parameters}

\begin{table}[t]
    \centering
    \begin{tabular}{|c|@{\hskip 0.05in}|c|c|c|c|c|}
        \hline
        Small-GAN & A & B & C & D & E \\
        \hline
        \textbf{16.73} & 18.75 & 18.09 & 17.03 & 17.88 & 17.45 \\
        \hline
    \end{tabular}
    \caption{FID scores for CIFAR using SN-GAN.
    The experiment list is: A = Training an SN-GAN, 
    B = Core-set selection directly on the images,
    C = Core-set applied directly on Inception embeddings without a random projection, 
    D = Core-set applied only on the prior distribution, 
    E = Core-set applied only on target distribution.}
    \label{tab:proposed_ablation_table}
\end{table}

We examine $i)$ the importance of the chosen target distribution 
for Core-set selection and $ii)$ the importance of performing Core-set on that target distribution.
The FID scores are reported in Table \ref{tab:proposed_ablation_table}. 

The importance of the target distribution is clear, since 
performing Core-set selection directly on the images (experiment B) performs similar to random-sampling. 
Experiment C supports our hypothesis that performing a random projection on the Inception embeddings can preserve semantic information while reducing the dimensionality 
of the features.
This increases the effectiveness of Core-set sampling and reduces sampling time.

Our ablation study also shows the importance of performing Core-set selection on both the 
prior and target distribution. 
The FID scores of the models are considerably worse when Core-set sampling is used on either distribution alone.

\subsection{Examination of Sampling Factors}

\begin{table}[t]
    \centering
    \begin{tabular}{|c|c|c|c|c|c|c|c|c|}
        \hline
        A & B & C & D & E & F & G & H & I\\
        \hline
        18.01 & 17.8 & 17.59 & 17.12 & 16.83 & \textbf{16.73} & 16.9 & 17.95 & 20.79\\
         \hline
    \end{tabular}
    \caption{FID scores for CIFAR using SN-GAN.
    Each of the experiment shows a different pair of over-sampling factors for the prior and target
    distributions.
    The factors are listed as: sampling factor for prior distribution $\times$ 
    sampling factor for target distribution.
    A = $2\times2$; B = $2\times4$;
    C = $4\times2$; D = $4\times4$; 
    E = $8\times4$; F = $4\times8$; 
    G = $8\times8$; H = $16 \times 16$; 
    I = $32 \times 32$}
    \label{tab:sampling_factor_table}
\end{table}

Another important hyper-parameter for training GANs using Core-set selection is the sampling factor.
In Table \ref{tab:sampling_factor_table} we varied the factors by which both the prior and the target distributions were over-sampled.
We see that using 4 for the sampling factor for the prior and 8 for the sampling factor
for the target distribution results in the best performance. 

\section{Related Work}

\subsection{Variance Reduction in GANs}
Researchers have proposed reducing variance in GAN training from an optimization perspective, by directly changing the way each of the networks are optimized. 
Some have proposed applying the extragradient method \citep{chavdarova2019reducing}, and 
others have proposed casting the \textit{minimax} two-player game as a variational-inequality problem \citep{gidel2018variational}.
\citet{biggan} recently proposed to reduce variance directly by using large mini-batch sizes.

\subsection{Stability in GAN Training}
Stabilizing GANs has been extensively studied theoretically.
Researchers have worked on improving the dynamics of the two 
player minimax game in a variety of ways \citep{nagarajan2017gradient, mescheder2018training, mescheder2018convergence, li2017towards, arora2017generalization}. 
Training instability has been linked to the architectural properties of GANs: especially to the discriminator \citep{sngan}. 
Proposed architectural stabilization techniques include using Convolutional 
Neural Networks (CNNs) \citep{dcgan}, using very large batch sizes \citep{biggan}, 
using an ensemble of the discriminators \citep{durugkar2016generative}, using spectral normalization for the discriminator \citep{sngan}, adding self-attention layers for the generator and discriminator networks \citep{attention, sagan} and using iterative updates to a \textit{global} generator and discriminator using an ensemble of paired generators 
and discriminators \citep{chavdarova2018sgan}. 
Different objectives have also been proposed to stabilize GAN training 
\citep{wgan, improved_wgan, mmdgan, lsgan, fishergan, cramergan}.

\subsection{Core-set Selection}
Core-set sampling has been widely studied from an algorithmic perspective in attempts to find better approximate solutions to the original NP-Hard problem 
\citep{agarwal2005geometric, clarkson2010coresets, pratap2018faster}. 
The optimality of the sub-sampled solutions have also been studied theoretically
\citep{barahona2005near, goldman1971optimal}. 
See \citet{phillips2016coresets} for a recent survey on Core-set selection algorithms. 
Core-sets have been applied to many machine learning problems such as $k$-means and approximate 
clustering \citep{har2004coresets, har2007smaller, badoiu2002approximate}), active learning for SVMs 
\citep{tsang2005core, tsang2007simpler}, unsupervised subset selection for hidden Markov models 
\citep{wei2013using} scalable Bayesian inference, \citep{huggins2016coresets} and mixture models \citep{feldman2011scalable}. 
We are not aware of Core-set selection being applied to GANs.

\subsection{Core-set Selection in Deep Learning}
Core-set selection is largely underexplored in the Deep Learning literature, but interest has recently increased. 
\citet{sener2017active} proposed to use Core-set sampling as a batch-mode active learning sampler for CNNs. 
Their method used the embeddings of a trained network to sample from. \citet{mussay2019activation} 
proposed using Core-set selection on the activations of a neural network for network compression. 
Core-set selection has also been used in continual learning to sample points for episodic memory \citep{vcl}.

\section{Conclusion}

In this work we present a general way to mimic using a large batch-size in GANs while minimizing computational overhead. 
This technique uses Core-set selection and improves performance in a wide variety of contexts.
This work also suggets further research: a similar method could be applied to other learning tasks where large mini-batches may be useful.

\section{Acknowledgements}
We thank Colin Raffel for feedback on a draft of this paper.

\bibliography{iclr2020_conference}
\bibliographystyle{iclr2020_conference}


\end{document}